\documentclass[11pt,a4paper]{article}
\usepackage[hyperref]{emnlp-ijcnlp-2019}
\usepackage{times}
\usepackage{latexsym, amsmath, amssymb}
\usepackage{graphicx}
\usepackage{soul}
\usepackage{placeins}
\usepackage{subcaption}
\usepackage{mwe}
\usepackage{pifont}
\usepackage{xcolor}

\usepackage{url}

\aclfinalcopy 

\title{Grounding learning of modifier dynamics:\\An application to color naming}

\author{Xudong Han$^\spadesuit$ \\ \And
  Philip Schulz $^\heartsuit$ \\
  $^\spadesuit$ University of Melbourne \\
  $^\heartsuit$ Amazon Research \\
  {\tt xudongh1@student.unimelb.edu.au} \\
  {\tt phschulz@amazon.com}\\
  {\tt trevor.cohn@unimelb.edu.au} \\\And
  Trevor Cohn $^\spadesuit$ \\}

\date{}

\begin{document}
\maketitle
\begin{abstract}
Grounding is crucial for natural language understanding. An important subtask is to understand modified color expressions, such as \emph{``dirty blue''}. We present a model of color modifiers that, compared with previous additive models in RGB space, learns more complex transformations. In addition, we present a model that operates in the HSV color space. We show that certain adjectives are better modeled in that space. To account for all modifiers, we train a hard ensemble model that selects a color space depending on the modifier-color pair. Experimental results show significant and consistent improvements compared to the state-of-the-art baseline model.\footnote{Code available at \url{https://github.com/HanXudong/GLoM}}
\end{abstract}

\section{Introduction}

Grounded color descriptions are employed to describe colors which are not covered by basic color terms~\cite{monroe2017colors}. For instance, \emph{``greenish blue"} cannot be expressed by only \emph{``blue"} or \emph{``green"}. Grounded learning of modifiers, as a result, is essential for grounded language understanding problems such as image captioning~\cite{Karpathy_2015_CVPR}, visual question answering~\cite{balanced_vqa_v2} and object recognition~\cite{5204091}.

\begin{figure}
    \centering
    \includegraphics[ scale = 0.3 ]{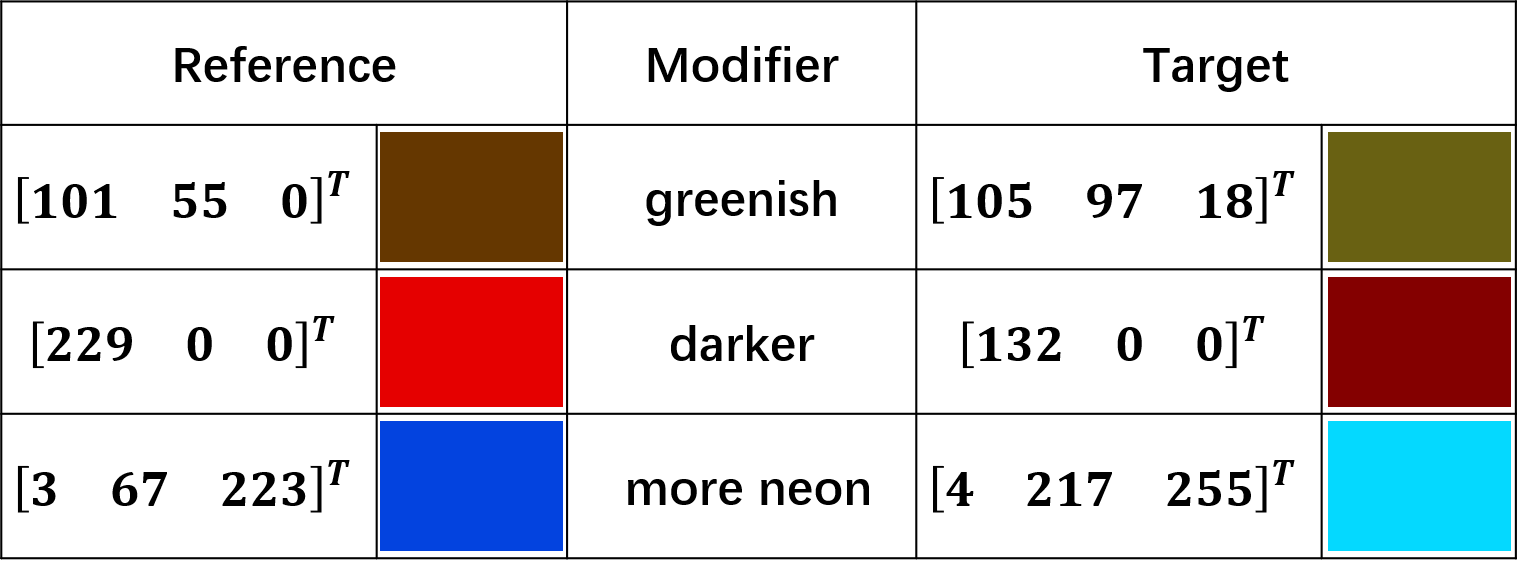}
    \caption{ Examples of the grounded modifier modelling task, shown in RGB space. Given the reference and modifier, the system must predict the target color.}
    \label{fig:tasks}
\end{figure}

In this paper, we present models that are able to predict the RGB code of a target color given a reference color and a modifier. For example, as shown in Figure~\ref{fig:tasks}, given a reference color code $ \vec{r} = \begin{bmatrix} 101 & 55 & 0 \end{bmatrix}^\top $ and a modifier $m=$~\emph{``greenish''}, our models are trained to predict the target color code $\vec{t} = \begin{bmatrix} 105 & 97 & 18 \end{bmatrix}^\top $. 
 The state-of-the-art approach for this task \cite{winn-muresan-2018-lighter} represents both colors and modifiers as vectors in RGB space, and learns a vector representation of modifiers $\vec{m}$ as part of a simple additive model, $\vec{r} + \vec{m} \approx \vec{r}$, in RGB color space.
 For instance, given the reference color $ \vec{r} = \begin{bmatrix} 229 & 0 & 0 \end{bmatrix}^\top$, the target color $ \vec{t} = \begin{bmatrix} 132 & 0 & 0 \end{bmatrix}^\top$, the modifier $m=$ \emph{``darker''} is learned as a vector $ \vec{m} = \begin{bmatrix} -97 & 0 & 0 \end{bmatrix}^\top$. 
 This model works well when the modifier is well represented as a single vector independent of the reference color,\footnote{\newcite{winn-muresan-2018-lighter} parameterize $\vec{m}$ as a function of $m$ and $\vec{r}$, such that the modifier vectors can adapt to the different reference colors, but showed that this had limited effect in practise.} but fails to model modifiers with more complex transformations, for example, color related modifiers, like \emph{``greenish''}, which are better modelled through color interpolation.


To fit a better model, we assume that there are approximate intersection points for the extension lines of modifier vectors, for instance, Figure~\ref{fig:dark_RGB} shows the \emph{``darker''} related vectors in RGB space and we can see that the intersection point is approximately $ \begin{bmatrix} 0 & 0 & 0 \end{bmatrix}^\top $. On the basis of this, we introduce an RGB model which can learn a transformation matrix and an interpolation point for each modifier.

There are many other color spaces besides RGB, e.g., HSL and HSV~\cite{HSV_1, HSV_2}, and mapping between such spaces can be done via invertible transformations~\cite{HSV2RGB}. As shown in Figure~\ref{fig:dark_HSV}, for some modifiers, color vectors can be approximately parallel in HSV space, thus simplifying the modeling problem. We propose a HSV model using a von-Mises loss, and show that this model outperforms the RGB method for many modifiers. We also present an ensemble model for color space selection, which determines for each modifier the best color space model. Overall our methods substantially outperform prior work, achieving state-of-the-art performance in the grounded color modelling task. 

\section{Methods}

 Here we describe the methods employed for color modeling. Formally, the task is to predict a target color vector $ \vec{t} $ in RGB space from a reference color
 vector $ \vec{r} $ and a modifier string $ m $. 

\subsection{Modeling in RGB space}

\paragraph{Baseline Model} \newcite{winn-muresan-2018-lighter} present a model (WM18) which represents a vector $ \vec{m} \in \mathbb{R}^{3} $ as a function of ($  m,\ \vec{r} $) pointing from a reference color vector $ \vec{r} $ to the target color vector $ \vec{t} $, such that $\vec{t} = \vec{r} + \vec{m} $. In the simplest case the modifier $ m $ is irrelevant to the reference color $ \vec{r} $. This assumption, however, does not hold in all situations. For example, when predicting an instance 
with the reference vector $ \vec{r} =  \begin{bmatrix} 193 & 169 & 106 \end{bmatrix}^\top $ and the modifier \emph{``greenish''}, the outcome $ \hat{t} $ is expected to be $ \begin{bmatrix} 177 & 183 & 102 \end{bmatrix}^\top $, however, WM18 predicts $ \hat{t} = \begin{bmatrix} 195 & 156 & 95 \end{bmatrix}^\top $. The cosine similarity between $(\hat{t} - \vec{r})$ and $(\vec{t} - \vec{r})$ is $ -0.76 $, i.e., $ \vec{m} $ points in the opposite direction to where it should.

\paragraph{RGB Model} 
As shown in Figure~\ref{fig:dark_RGB}, pairs of vectors for the same modifier are often not parallel. In theory, such vectors can even be orthogonal: compare \emph{``darker red''} vs \emph{``darker blue''}, which fall on different faces of the RGB cube. To model this, we propose a model in RGB space as follows:
\begin{equation}
    \label{eq:fullCovarianceRGB}
    \vec{t} = M\vec{r} + \vec{\beta}
\end{equation}
Where $ M \in \mathbb{R}^{3\times3} $ is a transformation matrix and $\vec{\beta}$ is a modifier vector which is designed to capture the information of $ m $. 
 Given an error term $\varepsilon \sim \mathcal{N}(\vec{0},\sigma I_{3})$,
the RGB model is trained to minimize the following loss for the log Gaussian likelihood:
\begin{equation}
    \label{eq:RGB_loss}
    \mathcal{L} = \frac{1}{n}\sum_{i=1}^{n}(\vec{t}_{i}-\hat{t_i})^\top(\vec{t}_{i}-\hat{t_i})
\end{equation}
where $\vec{t}_{i}$ is the target vector in each instance and $\hat{t}_{i}$ is the prediction.\footnote{Although other distributions, such as Beta distribution, may model uncertainties better, as focusing on mean value prediction, Gaussian models provide similar performances.}

\paragraph{Specific Settings} Our model generalizes WM18, which can be realized by setting $ M=I_{3}$ and $ \vec{\beta} = \vec{m} $.

Another interesting instance of the model is obtained by setting $M = (1-\alpha_m)I_3 $ and $ \vec{\beta} = \alpha_m\vec{m} $, which we call the Diagonal Covariance (DC) model. In contrast to the RGB model and WM18, which model $\vec{m}$ as a function of $ \vec{r} $ and $ m $, $ \vec{m} $ in the DC model does not depend on $ \vec{r} $. Given $\vec{r}$ and $m$, to predict the $ \vec{t} $, our DC model predicts $\vec{m}$ first and then applies a linear transformation to get the target color vector as follows: 
$\vec{t} = \vec{r} + \alpha_m\times(\vec{m}-\vec{r})$,
where $\alpha_m \in [0,1]$ is a scalar which only depends on $m$ and measures the distance from $ \vec{r} $ to $ \vec{m} $. In the DC model $ \vec{m} $ is the interpolation point for modifiers, such as $ \begin{bmatrix} 0 & 0 & 0 \end{bmatrix}^\top $ for the modifier \emph{``darker''}.

\subsection{Modeling in HSV space}

\begin{figure}
\centering
    \begin{subfigure}[b]{0.22\textwidth}
        \centering
        \includegraphics[width=\textwidth]{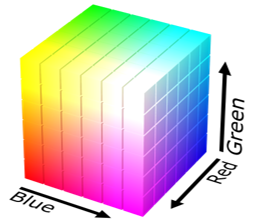}
        \caption[Network2]%
        {{\small RGB space}}    
        \label{fig:RGB_space}
    \end{subfigure}
    \hfill
    \begin{subfigure}[b]{0.22\textwidth}  
        \centering 
        \includegraphics[width=\textwidth]{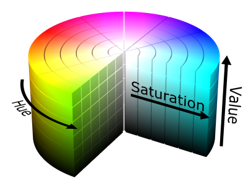}
        \caption[]%
        {{\small HSV space }}    
        \label{fig:HSV_space}
    \end{subfigure}
    \vskip\baselineskip
    \begin{subfigure}[b]{0.24\textwidth}   
        \centering 
        \includegraphics[width=\textwidth]{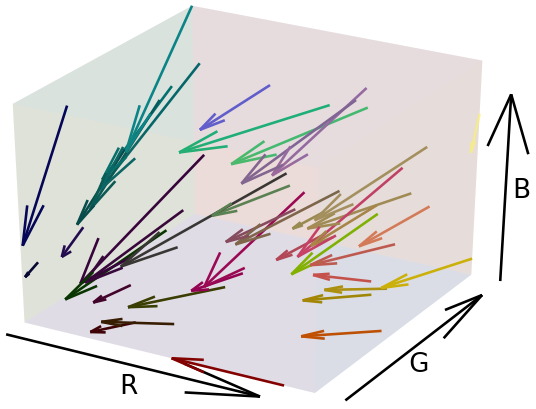}
        \caption[]%
        {{\small ``dark'' in RGB}}    
        \label{fig:dark_RGB}
    \end{subfigure}
    \quad
    \begin{subfigure}[b]{0.18\textwidth}   
        \centering 
        \includegraphics[width=\textwidth]{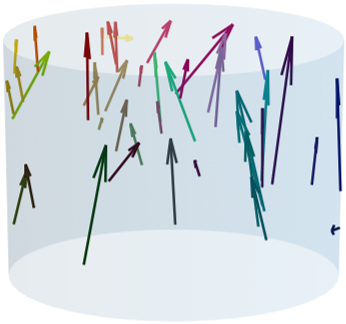}
        \caption[]%
        {{\small ``dark'' in HSV}}    
        \label{fig:dark_HSV}
    \end{subfigure}
\caption[ The average and standard deviation of critical parameters ]
{\small \ref{fig:RGB_space}: RGB color space. \ref{fig:HSV_space}: HSV color space. Arrows in \ref{fig:dark_RGB} and \ref{fig:dark_HSV} show vectors staring from reference colors to target colors in RGB and HSV color spaces. Images \ref{fig:RGB_space} and \ref{fig:HSV_space} by Michael Horvath, available under Creative Commons Attribution-Share Alike 3.0 Unported license. } 
\label{fig:examples_of_vec_m}
\end{figure}

Compared with the RGB color space, when modeling modifiers in HSV, there are two main differences: hue is represented as an angular dimension, and the modifier vectors are more frequently parallel (see Fig~\ref{fig:dark_HSV}). As shown in~Figure \ref{fig:HSV_space}, HSV space forms a cylindrical geometry with \emph{hue} as its angular dimension, with the value red occurring at both $0$ and $2\pi$. For this reason, modelling the hue with a Gaussian regression loss is not appropriate.
To account for the angularity, we model \emph{``hue''} with a von-Mises distribution, with the following \emph{pdf}:
\begin{equation}
    f(h) = \frac{\exp\left(k \cos{(h - \hat{h}})\right)}{2\pi I_0(k)} \, .
\end{equation}
The mean value $\hat{h}$ represents the center of \emph{hue} dimension, $k$ indicates the concentration about the mean, and $I_0(k)$ is the modified Bessel function of order 0.

When training the model, the parameter $k$ is assumed constant, and thus the loss function is:
\begin{equation}
    \label{eq:HSV_loss}
    \mathcal{L} = 1 - \frac{1}{n}\sum_{i=1}^n \cos{(h_i - \hat{h}_i)} \, ,
\end{equation}
where $h_i$ is the \emph{hue} value of the target color in each instance and $\hat{h}_i$ is the prediction. 

The second difference to modeling in RGB space is that the modifier behavior is simpler in HSV. For modifiers, vectors from reference colors to target colors are more likely to be parallel (see Figure~\ref{fig:dark_HSV}). As a result, we present an additive model in HSV space where a modifier $m$ will be modeled as a vector from $\vec{r}$ to $\vec{t}$:
\begin{equation}
    \label{eq:HSVmodel}
    \vec{t} = \vec{r} + \vec{m}
\end{equation}
Here $\vec{m}$ is a function of both $ m $ and $ \vec{r} $. In addition, modifier modeling will be split into two parts: modeling \emph{``hue''} dimension as von-Mises distribution and other dimensions together as a bivariate normal distribution (See Equation~\ref{eq:RGB_loss}). Notice that Equation~\eqref{eq:HSVmodel}
is the same equation as used by WM18, however, here it is applied in a color space that better fits its assumptions.

 WM18 is presented and evaluated only in RGB color space. To compare its performance with our models, we transform output into RGB space.

\subsection{Ensemble model}
An ensemble model is trained to make the final prediction, which we frame as a hard binary choice to select the color space appropriate for the given modifier.
This works by applying the general RGB model and HSV model (Equations~\ref{eq:fullCovarianceRGB} and~\ref{eq:HSVmodel}) to get their predictions, and converting the HSV predictions into RGB space. Then the hard ensemble is trained to predict which color space should be used based on the modifier $m$ and the reference vector $ \vec{r} $, using as the learning signal which model prediction had the smallest error against the reference colour for each instance (measured using Delta-E distance, see \S\ref{experiments:eval}). The probability of the RGB model being selected is: $ p = \sigma(f(m,\ \vec{r})) $, where $ \sigma $ is the logistic sigmoid function and $ f(m,\ \vec{r}) $ is a function of modifier $ m $ and reference color $ \vec{r} $.


\section{Experiments} \label{experiments}

\subsection{Dataset}

The dataset\footnote{\url{https://bitbucket.org/o_winn/comparative_colors}} used to train and evaluate our model includes 415 triples (reference color label, $r$, modifier, $m$, and target color label, $t$) in RGB space presented by \newcite{winn-muresan-2018-lighter}. \newcite{Color-Survey} collected the original dataset consisting of color description pairs collected in an open online survey; the dataset was subsequently filtered by \newcite{FliteredXKCD}. Winn and Muresan processed color labels and converted pairs to triples with 79 unique reference color labels and 81 unique modifiers.

We train models in both RGB and HSV color space, but samples in WM18 are only presented in RGB space. Because modifiers encode the general relationship between $r$ and $t$ we use the same approach presented by~\newcite{winn-muresan-2018-lighter}: using the mean value of a set of points to represent a color.
A drawback of this approach is that it does not account for our uncertainty about the appropriate RGB encoding for a given color word.

\begin{table*}[h!t]
    \centering
\begin{tabular}{c|ccccccc}
                   & \multicolumn{5}{c}{\textbf{Cosine Similarity $\pm$ SD } ($\uparrow$)}           \\ \hline
Test Condition     & {\bf RGB}        & {\bf WM18}   & {\bf HSV}            & {\bf Ensemble}           & {\bf WM18$^*$}\\
Seen Pairings      & 0.954$\pm$0.001    & 0.953$\pm$0.000   & 0.934$\pm$0.089   & \textbf{0.954}$\pm$0.0    & 0.68 \\
Unseen Pairings    & 0.799$\pm$0.044    & 0.771$\pm$0.032   & \textbf{0.843}$\pm$0.144   & 0.797$\pm$0.0    & 0.68 \\
Unseen Ref. Color  & 0.781$\pm$0.015    & 0.767$\pm$0.010   & \textbf{0.945}$\pm$0.019   & 0.804$\pm$0.0    & 0.40 \\
Unseen Modifiers    & 0.633$\pm$0.042    & 0.637$\pm$0.032   & \textbf{0.724}$\pm$0.131   & 0.629$\pm$0.0 & 0.41          \\
Fully Unseen       & 0.370$\pm$0.029    & 0.358$\pm$0.038   & \textbf{0.919}$\pm$0.026   & 0.445$\pm$0.0 & -0.21      \\
Overall            & 0.858$\pm$0.006    & 0.856$\pm$0.003   & \textbf{0.911}$\pm$0.057   & 0.868$\pm$0.0    & 0.65 \\ \hline
                   & \multicolumn{5}{c}{\textbf{Delta-E  $\pm$ SD  } ($\downarrow$)}           \\ \hline
Test Condition      & RGB               & WM18              & HSV               & Ensemble              & WM18$^*$\\
Seen Pairings       & \textbf{3.121}$\pm$0.027   & 3.188$\pm$0.062   & 5.380$\pm$4.846   & 4.093$\pm$0.1     & 6.1 \\
Unseen Pairings     & 6.454$\pm$0.233   & 6.825$\pm$0.093   & 11.701$\pm$3.358  & \textbf{5.873}$\pm$0.0     & 7.9 \\
Unseen Ref. Color   & 7.456$\pm$0.184   & 7.658$\pm$0.363   & 10.429$\pm$2.523  & \textbf{7.171}$\pm$0.0     & 11.4 \\
Unseen Modifiers    & 13.288$\pm$1.082  & 13.891$\pm$1.077  & 14.183$\pm$5.175  & 10.927$\pm$0.0    & \textbf{10.5} \\
Fully Unseen        & 13.859$\pm$0.874  & 14.516$\pm$0.587  & \textbf{12.432}$\pm$2.170  & 13.448$\pm$0.0    & 15.9 \\
Overall             & \textbf{5.412}$\pm$0.169   & 5.595$\pm$0.128   & 7.487$\pm$3.940   & 5.777$\pm$0.0     & 6.8 \\

\end{tabular}
    \caption{Average cosine similarity score and Delta-E distance over 5 runs. A smaller Delta-E distance means a less significant difference between two colors. \textbf{Bold}: best performance. Hard: the hard ensemble model. WM18$^*$: the performance from  WM18 paper. See \emph{Supplementary Material} for example outputs and ensemble analysis.}
    \label{tab:performance}
\end{table*}

\subsection{Experiment Setup}
\label{experiments:eval}

\paragraph{Model configuration:} The model presented by \newcite{winn-muresan-2018-lighter} is initialized with Google's pretrained 300-d word2vec embeddings \cite{w2v-a, DBLP:journals/corr/abs-1301-3781} which are not updated during training. To perform comparable experiments, all models in paper are designed with the same pre-trained embedding model. Other pretrained word embeddings, such as GloVe \cite{pennington2014glove} and BERT \cite{devlin2019bert}, were also tested but there was no significant difference in performance compared to word2vec. Single models are trained over 2000 epochs with batch size 32 and 0.1 learning rate. The hyper-parameters for the ensemble model are as follows: 600 epochs, 32 batch size, and 0.1 learning rate.

\paragraph{Architecture:} An input modifier is represented as a vector by word2vec pretrained embeddings and followed by two fully connected layers($FC_1$ and $FC_2$) with size 32 and 16 respectively. Let $h_1$ be the hidden state of $ FC_2 $ then $h_1 = FC_2(FC_1(\vec{r}, \ E_m,\ \vec{r})$ where $ E $ are fixed, pretrained word2vec embeddings. $ \vec{r} $ is used as an input for both $ FC_1 $ and $ FC_2 $. After $ FC_2 $, all the other layers are based on hidden state $ h_1 $.

\paragraph{Evaluation:} Following \newcite{winn-muresan-2018-lighter}, we evaluate the performance in 5 distinct input conditions: (1) \emph{Seen Pairings:} The triple $(r, m, t)$ has been seen when training models. (2) \emph{Unseen Pairings:} Both $r$ and $m$ have been seen in training data, but not the triple $(r, m, t)$. (3) \emph{Unseen Ref. Color:} $r$ has not been seen in training, while $m$ has been seen. (4) \emph{Unseen modifiers:}  $m$ has not been seen in training, while $r$ has been seen. (5) \emph{Fully Unseen:} Neither $r$ nor $m$ have been seen in training.

Because of the small size of the dataset, we report the average performance over 5 runs with different random seeds. Two scores, cosine similarity, and Delta-E are applied for evaluating the performance. Cosine similarity measures the difference in terms of vector direction in color space and Delta-E is a non-uniformity metric for measuring color differences. Delta-E was first presented as the Euclidean Distance in CIELAB color space \cite{mclaren1976xiii}. Lower Delta-E values are thus preferable as they indicate better matching of the target color. \newcite{luo2001development} present the latest and most accurate CIE color difference metrics, Delta-E 2000, which improve the original formula by taking into account weighting factors and fixing the lightness inaccuracies. Our models are evaluated with Delta-E 2000.

\subsection{Results}

Table \ref{tab:performance} shows the results. 
Compared with WM18, our RGB model outperforms under all conditions. As we have stated, our model is a generalization of their approach. The more complex transformation matrix in our RGB model is able to learn more information, such as the effects of covariance between color channels, and thus achieves a better performance than WM18.
Note that our reimplementation of the original WM18 system lead to significantly better performance.\footnote{We train the model for many more epochs, which is required for a good to the training data. This can be observed in terms of the `seen pairings', where the WM18 reported results appear to be underfitting.}

According to the cosine similarity, the HSV model is superior for most test conditions (confirming our hypothesis about simpler modifier behaviour in this space). However for Delta-E, the RGB model and ensemble perform better.
Unlike cosine, Delta-E is sensitive to differences in vector length, and we would argue it is the most appropriate metric because lengths are critical to measuring the extent of lightness and darkness of colors.
Accordingly the HSV model does worse under this metric, as it more directly models the direction of color modifiers, but as a consequence this leads to errors in its length predictions.
Overall the ensemble does well according to both metrics, and has the best performance for several test conditions with Delta-E.

\begin{figure}
    \centering
    \includegraphics[ scale = 0.491 ]{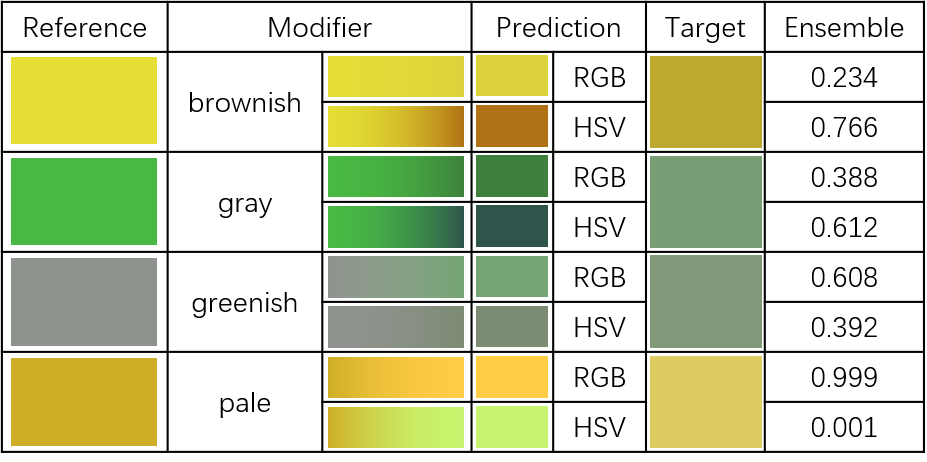}
    \caption{Examples of predictions from RGB, HSV and Ensemble model. The Ensemble column reports the predicted probability of the RGB and HSV models being selected.}
    \label{fig:HSV_TH}
\end{figure}

\paragraph{Error Analysis} We first focused error analysis on prediction of ``Unseen Modifiers'' and ``Fully Unseen'' instances. As shown in Table \ref{tab:performance}, our models are able to predict target colors given seen modifiers but fail to make predictions for instances with unseen modifiers. All modifiers are represented by word2vec embeddings, and we expect that predictions of unseen modifiers should be close to instances with similar seen modifiers. For example, the prediction of a reference color $\vec{r}$ and modifier \emph{``greeny''} should be similar to the prediction of the same reference color $\vec{r}$ and a similar seen modifier, e.g. \emph{``green''} and \emph{``greenish''}. However, the prediction of \emph{``greeny''} is more similar to \emph{``bluey''}, a consequence of these terms having highly similar word embeddings (as do other colour modifiers with a \emph{-y} suffix, irrespective of their colour). This is related to the problem reported in
\newcite{mrkvsic2016counter}, whereby words and their antonyms often have similar embeddings, as a result of sharing similar distributional contexts.
Accordingly for the \emph{unseen modifier} condition, our model is often misled by attempting to generalise from nearest neighbour modifiers which have a different meaning.

\section{Related Work}

\newcite{baroni2010nouns} was the first work to propose an approach to adjective-noun composition (AN) for corpus-based distributional semantics which represents nouns as vectors and adjectives as matrices nominal vectors. However it is hard to gain an intuition for what the transformation does since these embeddings generally live on a highly structured but unknown manifold. In our case, we operate on colors and we actually know the geometry of the colour spaces we use. This makes it easier for us to interpret the learned mapping (see Figure \ref{fig:dark_RGB} and \ref{fig:dark_HSV} that show convergence to a point in RGB and parallelism in HSV space). 

\section{Conclusion and Future Work}

In this paper, we proposed novel models of predicting color based on textual modifiers, incorporating a matrix transformation than the previous largely linear additive method. As well as our more general approach, we exploit the properties of another color space, namely HSV, in which the modifier behaviours are often simpler. Overall our method leads to state of the art performance on a standard dataset.


In future work, we intend to develop more accurate modifier representations to allow for better generalisation to unseen modifiers. This might be achieved by using a composition sub-word representation for modifiers, such as character-level encoding.
%
%
Finally, we also strive to acquire larger datasets. This is a crucial step towards comparing the generalization performance of different color-modifier models. Models trained on larger data sets are likely to be more applicable to real-world problems since they learn representations for more color terms.

\bibliography{emnlp-ijcnlp-2019}
\bibliographystyle{acl_natbib}


\end{document}